# Onboard Wind Estimation for Small UAVs Equipped with Low-Cost Sensors: An Aerodynamic Model-Integrated Filtering Approach

**BINGCHEN CHENG**

Shenyuan Honors College, Beihang University, Beijing, 100191, China

**TIELIN MA**

**JINGCHENG FU**

**LULU TAO**

Institute of Unmanned System, Beihang University, Beijing, 100191, China

**TIANHUI GUO**

School of Aeronautical Science and Engineering, Beihang University, Beijing, 100191, China.

Manuscript received XXXXX 00, 0000; revised XXXXX 00, 0000; accepted XXXXX 00, 0000.

This work was supported by Beijing Natural Science Foundation (No. QY24131), the Fundamental Research Funds for the Central Universities (No. 501QYJC2024129001), Aeronautical Science Foundation of China (No. 2024Z006051002), and the Taihu Lake Innovation Fund for Future Technology (No. 1311402).

Authors' address: B. Cheng is now with the Shenyuan Honors College, Beihang University, Beijing 100191, China (e-mail: 21375038@buaa.edu.cn); T. Ma, J. Fu, and L. Tao are now with the Institute of Unmanned System, Beihang University, Beijing 100191 (e-mail: matielin@buaa.edu.cn, fujingcheng@buaa.edu.cn, 21374241@buaa.edu.cn); T. Guo is now with the School of Aeronautical Science and Engineering, Beihang University, Beijing 100191, China (e-mail: tianhuiguo@buaa.edu.cn). (Corresponding author: Jingcheng FU).

Mentions of supplemental materials and animal/human rights statements can be included here.

Color versions of one or more of the figures in this article are available online at http://ieeexplore.ieee.org.

*Abstract*—To enable autonomous wind estimation for energy-efficient flight in small unmanned aerial vehicles (UAVs), this study proposes a method that estimates flight states and wind using only the low-cost essential onboard sensors required for autonomous flight, without relying on additional wind measurement devices. The core of the method includes an Extended Kalman Filter (EKF) integrated with the aerodynamic model and an Adaptive Moving Average Estimation (AMAE) technique, which improves the accuracy and smoothness of the wind estimation. Simulation results show that the approach efficiently estimates both steady and time-varying 3D wind vectors without requiring flow angle measurements. The impact of aerodynamic model accuracy on wind estimation errors is also analyzed to assess practical applicability. Flight tests validate the effectiveness of the method and its feasibility for real-time onboard computation. Additionally, uncertainties and error sources encountered during testing are systematically examined, providing a foundation for further refinement.

## I. INTRODUCTION

Limited by size and cost, small unmanned aerial vehicles (UAVs) have restricted battery or fuel weight, which leads to constraints in endurance and mission capabilities. Enhancing the endurance performance of small UAVs can enable long-duration missions and broaden application scenarios. In nature, the flight strategies of birds provide crucial inspiration for energy-efficient flight in small UAVs [1]. For instance, albatrosses, falcons, and migratory birds can autonomously perceive specific wind in the environment [2], such as gradient winds, thermal updrafts, and favorable wake flows. They subsequently plan and execute strategies like dynamic soaring, static gliding, and multi-agent formation flight, effectively utilizing wind energy to reduce flight energy consumption [3]. These strategies enable birds to demonstrate remarkable endurance during transoceanic migrations [4], which fundamentally arises from their precise perception of the spatiotemporal distribution of three-dimensional wind and energy utilization mechanisms, offering valuable insights for improving UAV endurance [5]. Since Rayleigh [6] discovered this phenomenon and established the classical model of dynamic soaring in 1883, research on bio-inspired energy-efficient flight strategies for UAVs—such as dynamic soaring [7], static gliding [8], and multi-UAV formation flight [9]—has deepened continuously, gradually forming several core technical directions [10]. These include energy-optimal flight trajectory planning based on optimal control theory, robust trajectory tracking control for complex flight paths, and research on UAV morphing technologies adapted to multi-modal wind. The effective implementation of these studies relies heavily on accurate estimation of wind in the flight environment [11]. For example, Hong et al. [12] proposed using real-time wind information to refine dynamic soaring trajectories, underscoring the critical importance of real-time wind estimation for energy-efficient flight.

Current wind estimation techniques exhibit significant differentiation in their application-specific characteristics. In meteorological research, high-precision wind measurement instruments—such as ultrasonic anemometers, Doppler lidars, and radiosondes—are commonly employed alongside advanced computational methods. These methods include computational fluid dynamics (CFD) [13], deep learning prediction models [14], swarm intelligence techniques [15], and multi-feature fusion probabilistic prediction [16] for wind estimation. However, the wind field estimated in this domain typically provides significant spatiotemporal-scale estimates, which struggle to meet the fine-grained wind perception requirements of small UAVs during energy-efficient flight operations such as dynamic soaring, static gliding, and multi-UAV formation flight. Furthermore, the high computational demand impedes real-time implementation. Consequently, achieving autonomous wind estimation using onboard sensors becomes critical.

For autonomous wind estimation in UAVs, some studies have explored dedicated aircraft platform designs. For instance, special delta-wing aerodynamic configurations [17] sense wind disturbances to estimate wind speed. While demonstrating good estimation performance, the versatility and multi-mission capabilities of such platforms are often limited. Concurrently, other research has begun investigating novel sensor schemes, such as utilizing wing pressure sensor arrays [18] to replace conventional airspeed and flow angle measurements. However, deployment convenience poses a challenge for small UAVs. Fusing visual information with wind observers [19] presents a novel approach for wind perception in ultra-lightweight UAVs, but imposes stringent requirements for specialized precision sensor design. Airborne lidar represents a sophisticated, autonomous wind-sensing technology for aircraft. Some studies have employed empirical Bayesian estimation methods [20] and physics-informed deep learning models [21] to enhance the capability of lidar for wind measurement accuracy and achieve 3D spatiotemporal wind field reconstruction. Nevertheless, constrained by the size and cost of small UAVs, such high-cost, computationally intensive precision sensors are unsuitable for real-time onboard computation on small UAVs.

To enhance wind perception capabilities for small UAVs, utilizing essential low-cost onboard sensors required for autonomous flight—including Global Navigation Satellite System (GNSS), Inertial Navigation System (INS), and Air Data System (ADS)—for autonomous wind estimation presents a more feasible approach. Based on UAV kinematics, a commonly employed method [22][23][24] solves the vector triangle relationship formed by airspeed, groundspeed, and wind speed. While its principle is clear, practical applications often require simplified handling of sensor noise, and its accuracy heavily relies on the precision of airspeed and groundspeed measurements. Optimization methods, such as the Generalized Model Predictive Static Programming (GMPSP) [25], demonstrate high accuracy and rapid response advantages in wind estimation simulations. Further research could investigate its performance in actual flight tests. Various filtering algorithms are widely adopted due to their strengths in data fusion and state estimation. Related studies employ Extended Kalman Filter (EKF) [26][27] to fuse data from sensors like GPS and pitot tubes, combined with specific flight maneuvering strategies, for wind estimation. However, simplifications in the motion model limit its effectiveness for estimating time-varying wind. The Ensemble Kalman Filter (EnKF) [28] can be used to detect specific wind structures, such as wake vortices; however, its robustness in highly turbulent environments requires further validation. The Unscented Kalman Filter (UKF) can jointly estimate attitude angles and wind speed by fusing GPS and INS data [29] or by incorporating simplified aerodynamic models [30]. Yet, it often faces challenges such as significant estimation errors due to model simplifications, high computational complexity, and reliance on high-precision Inertial Measurement Units (IMUs). Moving Horizon Estimation (MHE) [31] integrates multi-sensor data with an aerodynamic model for joint estimation, thereby improving accuracy under low-quality sensor configurations. However, it suffers from high computational load, inadequate lateral dynamics modeling, and crucially, the impact of aerodynamic model accuracy on estimation results has not been sufficiently analyzed or validated.

Synthesizing the above research status, for autonomous wind perception technologies that enable wind energy utilization in energy-efficient flight of small UAVs, there is an urgent need to achieve high-precision, real-time autonomous estimation of both steady and time-varying wind, including gust disturbances, under the constraints of utilizing only essential, low-cost onboard sensors for UAVs. This must effectively handle sensor noise and accommodate scenarios involving partial sensor failure or absence. Concurrently, analyzing the impact of aerodynamic model accuracy on wind estimation errors is of crucial importance for evaluating the practical application potential of this technology on small UAV platforms.

Therefore, this study addresses the wind estimation requirements for small UAVs by proposing a hybrid approach integrating an aerodynamic model-based EKF with Adaptive Moving Average Estimation (AMAE). This method utilizes information from onboard sensors such as GNSS, INS, and ADS to perform joint estimation of flight states and wind velocity vectors through EKF-based multi-source data fusion. Compared to methods like UKF, EnKF, and MHE, EKF offers greater advantages in meeting the demands of real-time onboard computation. By incorporating the UAV aerodynamic model into the EKF framework, the variations in aerodynamic forces and moments acting on the UAV can be precisely described. This enables practical estimation of both steady and time-varying wind, satisfying the requirements for real-time autonomous perception and dynamic updates of diverse specialized wind during energy-efficient flight. Furthermore, to enhance the smoothness and noise resistance of the estimation results, this study designs the AMAE for post-processing the wind estimates by adaptively adjusting the weights within a moving window, effectively suppressing noise interference while maintaining the capability to estimate time-varying wind.

The structure of the subsequent sections in this paper is organized as follows: Section II introduces the employed UAV platform and sensor characteristics, develops the UAV dynamic and aerodynamic models, and presents the state estimation methods along with the simulated wind environments. Section III analyzes the simulation results under various simulated wind conditions, based on the developed models and techniques, with a focus on investigating the impact of aerodynamic model accuracy on wind estimation errors. Section IV describes the flight test conditions and presents the experimental results. Finally, Section V summarizes the main conclusions of this study and outlines future research directions.

## II. MODELING AND METHODS

### A. Flight Platform and Sensor Characteristics

This study employs a practical and reliable small UAV with a flying-wing configuration. Its primary characteristic parameters are listed in Table I.

Fig. 1. illustrates the external features of the UAV and the placement locations of its sensors. To fully identify flight information such as speed and attitude, this study equips the UAV with a GNSS for ground speed, an INS for attitude angles, angular rates, and accelerations, and an ADS for airspeed. It is important to note that to accommodate scenarios where flow angle sensors within the ADS fail or where installing flow angle sensors is unsuitable for small UAVs, this study does not incorporate sensors to obtain observations of the angle of attack and sideslip angle during wind estimation.

**TABLE I**
**Primary Characteristic Parameters of the UAV**

| Parameter | Value |
|---|---|
| $m$ | 4.00 kg |
| $S$ | 0.75 m$^2$ |
| $b$ | 2.10 m |
| $\bar{c}$ | 0.42 m |

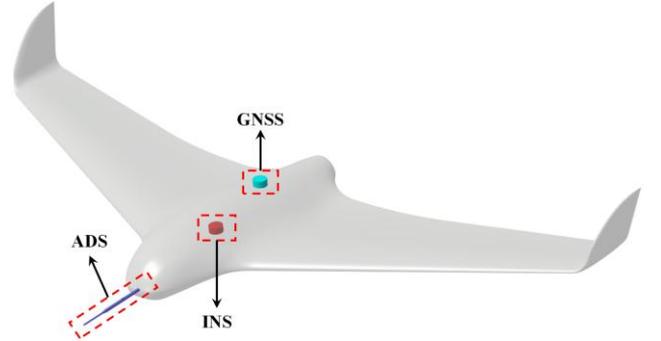

Fig. 1. Flight platform and sensor configuration.

To simulate real-world sensor noise characteristics, this study incorporates Gaussian white noise into the simulated sensor data. The simulated measurement accuracies are approximately $0.5°/s$ for INS angular rates, $1°$ for INS attitude angles, $0.1$ m/s$^2$ for INS accelerations, $0.3$ m/s for GNSS groundspeed, and $1$ m/s for ADS airspeed.

To assess the feasibility of real-time computation, sampling frequencies are configured at 100 Hz for INS, 100 Hz for ADS, and 10 Hz for GNSS.

### B. Dynamic and Aerodynamic Modeling

Dynamics modeling of the UAV assumes a rigid body with constant mass, constant gravitational acceleration, and a flat Earth model neglecting curvature. The definitions of the coordinate systems, along with relevant angles, velocities, and other parameters, are illustrated in Fig. 2.

A commonly used method for estimating wind is the velocity vector triangle approach. As shown in Fig. 3., this method calculates wind speed by establishing a velocity vector triangle using airspeed and groundspeed. According to the velocity vector triangle relationship, wind speed, ground speed, and airspeed satisfy (1) in the inertial coordinate system:

$$V_g = R_{gb}V_b + V_{wg} \tag{1}$$

where $V_g = [V_x, V_y, V_z]^T$ denotes ground speed, $V_b = [u, v, w]^T$ denotes airspeed, $V_{wg} = [w_x, w_y, w_z]^T$

denotes the wind speed, and $R_{gb}$ denotes the transformation matrix from the body frame to the inertial frame.

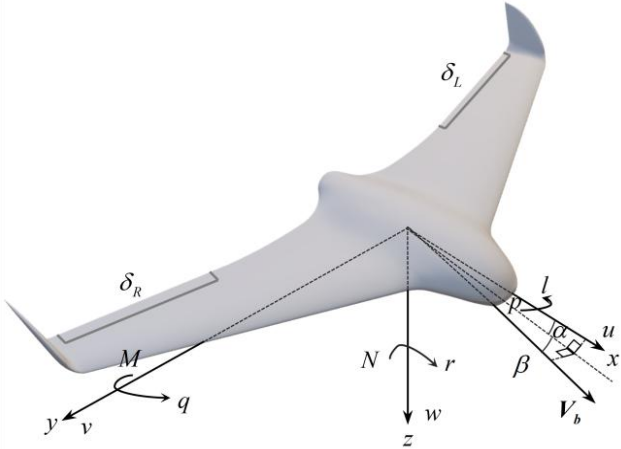

Fig. 2. Coordinate system definition.

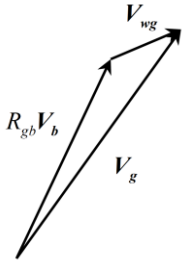

Fig. 3. Velocity vector triangle.

To calculate the wind speed, it is necessary to obtain additional information, including airspeed, angular rates, and attitude angles. According to Newton's second law, the equations governing the aircraft's translational and rotational motion yield:

$$\begin{aligned} m\dot{V}_b &= T[\cos\varphi \quad 0 \quad -\sin\varphi]^T + R_{ba}F_{aero} \\ &\quad + R_{bg}[0 \quad 0 \quad mg]^T - \omega \times (mV_b) \\ J\dot{\omega} &= M - \omega \times (J\omega) \\ \dot{\eta} &= T(\eta)\omega \end{aligned} \quad (2)$$

where $T$ denotes engine thrust, $\varphi$ denotes engine installation angle, $\omega = [p,q,r]^T$ denote angular rate components in the body frame, $\eta = [\psi,\theta,\phi]^T$ denote yaw angle, pitch angle, and roll angle, $F_{aero} = [D,C,L]^T$ denote drag force, side force, and lift force respectively, $M = [l,M,N]^T$ denote roll moment, pitch moment, and yaw moment respectively, $R_{ba}$ and $R_{bg}$ denote the transformation matrices from the wind frame to the body frame and from the inertial frame to the body frame respectively, $T(\eta)$ denotes the transformation matrix relating Euler angles to angular rates, $m$ denotes the UAV mass, $g$ denotes gravitational acceleration, and $J$ denotes the inertia matrix.

This study assumes $\varphi = 0°$, and engine thrust depends solely on throttle position $\delta_p$, satisfying:

$$T = T_p \cdot \delta_p. \quad (3)$$

Aerodynamic forces and moments satisfy:

$$[D \quad C \quad L]^T = \bar{q}S[C_D \quad C_C \quad C_L]^T \quad (4)$$

$$[l \quad M \quad N]^T = \bar{q}S[bC_l \quad \bar{c}C_m \quad bC_n]^T \quad (5)$$

where $\bar{q} = \frac{1}{2}\rho V^2$ denotes dynamic pressure, $\rho$ denotes atmospheric density, $V$ denotes airspeed, satisfying:

$$V = \sqrt{u^2 + v^2 + w^2} \quad (6)$$

$S$ denotes wing area, $b$ denotes wingspan, $\bar{c}$ denotes mac, and $[C_D, C_C, C_L, C_l, C_m, C_n]$ denote the drag coefficient, side-force coefficient, lift coefficient, roll moment coefficient, pitch moment coefficient, and yaw moment coefficient, governed by [32]:

$$\begin{aligned} C_L &= C_L(\alpha, q, \delta_e) \\ C_D &= C_D(\alpha, \delta_e) \\ C_C &= C_C(\beta, p, r, \delta_a) \\ C_l &= C_l(\beta, p, r, \delta_a) \\ C_m &= C_m(\alpha, q, \delta_e) \\ C_n &= C_n(\beta, p, r, \delta_a) \end{aligned} \quad (7)$$

where $[\alpha, \beta]$ denote the angle of attack and sideslip angle, $[\delta_a, \delta_e]$ denote the equivalent aileron deflection and elevator deflection. For the flying-wing UAV employed in this study, elevon surfaces are installed at the trailing edge of the wing. The equivalent aileron and elevator deflection angles are calculated from the deflection angles of the left and right control surfaces:

$$[\delta_a \quad \delta_e]^T = \left[\frac{\delta_R - \delta_L}{2} \quad \frac{\delta_R + \delta_L}{2}\right]^T. \quad (8)$$

$\alpha$、$\beta$ can be measured by the ADS, and its three-axis components in the body frame satisfy the relationship with airspeed:

$$\alpha = \arctan\left(\frac{w}{u}\right) \quad (9)$$

$$\beta = \arcsin\left(\frac{v}{V}\right). \quad (10)$$

### C. State Estimation

Wind estimation is closely integrated with UAV flight state estimation, necessitating the use of multi-source sensor information for joint estimation of both flight state and wind. Filtering algorithms serve as a means for multi-source data fusion [33], enabling the joint estimation and correction of low-confidence data from diverse sources to obtain high-confidence state estimates. These algorithms demonstrate robust estimation performance when handling noise-corrupted data sources. EKF is a classical and efficient filtering method applicable to nonlinear state estimation. It approximates the original nonlinear system model through first-order Taylor expansion, computes predicted state values in the state prediction step, and calculates the Kalman gain during the observation update step. This process optimally weights the predicted state values and observed measurements to yield high-confidence state estimation results. The technique is widely adopted due to its computational simplicity.

To establish the EKF estimation model for flight state and wind speed, the state prediction equation and observation update equation are first formulated:

By incorporating the 6-DOF dynamic equations, the UAV flight states $u, v, w, p, q, r, \psi, \theta, \phi$ are selected as state variables. Simultaneously, to estimate wind, $w_x, w_y, w_z$ are chosen as state variables under the assumption of slowly varying wind. Given that wind variation frequencies are significantly lower than sensor sampling frequencies, it can be reasonably approximated that wind remains essentially constant within the sampling interval, i.e., the wind speed rate satisfies:

$$\left[\dot{w}_x, \dot{w}_y, \dot{w}_z\right]^T = \mathbf{0}. \quad (11)$$

Thus, the state vector $\mathbf{x} = [u, v, w, p, q, r, \psi, \theta, \phi, w_x, w_y, w_z]^T$ is obtained. Due to this selection of state variables, wind speed can be estimated while simultaneously enabling the combinatorial estimation of airspeed, angle of attack, and sideslip angle using state variables through (6), (9), and (10). Furthermore, the aerodynamic forces and moments coefficients can be estimated by incorporating the aerodynamic model (7).

The equivalent aileron deflection, elevator deflection, and throttle position constitute the control vector $\mathbf{u} = [\delta_a, \delta_e, \delta_p]^T$.

Consequently, by incorporating (2)~(5), (7), and (11), the control vector and state vector satisfy the dynamic equation:

$$\dot{\mathbf{x}} = f(\mathbf{x}, \mathbf{u}). \quad (12)$$

The solution to this ODE (12) is obtained through numerical integration using the forward Euler method, with a discretization time step $\Delta t$. Introducing process noise $\mathbf{w}_{k-1}$ yields the state prediction equation:

$$\mathbf{x}_k = \mathbf{x}_{k-1} + \Delta t \cdot f(\mathbf{x}_{k-1}, \mathbf{u}_{k-1}) + \mathbf{w}_{k-1} \quad (13)$$

$\mathbf{w}_{k-1}$ follows a Gaussian distribution with mean 0 and variance $Q_{k-1}$, denoted as $\mathbf{w}_{k-1} \sim N(0, Q_{k-1})$.

When the UAV is not equipped with flow angle sensors, information acquired from GNSS, INS, and ADS is selected as observations, i.e. $z = [V_x, V_y, V_z, p, q, r, \psi, \theta, \phi, a_x, a_y, a_z, V]$. The observation equation is established using (1)~(8). Angular rates and attitude angles are directly obtained from state estimates $[p, q, r, \psi, \theta, \phi]^T = [p, q, r, \psi, \theta, \phi]^T$, while groundspeed, acceleration $\mathbf{a} = [a_x, a_y, a_z]$, and airspeed are computed as:

$$\begin{aligned} \mathbf{V}_g &= R_{gb}\mathbf{V}_b + \mathbf{V}_{wg} \\ \mathbf{a} &= \left(T[\cos\varphi \quad 0 \quad -\sin\varphi]^T + R_{ba}\mathbf{F}_{aero}\right)/m \\ &\quad + R_{bg}[0 \quad 0 \quad g]^T \\ V &= \sqrt{u^2 + v^2 + w^2}. \end{aligned} \quad (14)$$

The observation equation is expressed as:

$$z = h(\mathbf{x}, \mathbf{u}). \quad (15)$$

After discretization and introduction of observation noise $\mathbf{v}_k$:

$$z_k = h(\mathbf{x}_k, \mathbf{u}_k) + \mathbf{v}_k \quad (16)$$

$\mathbf{v}_k$ follows a Gaussian distribution with mean 0 and variance $R_k$, denoted as $\mathbf{v}_k \sim N(0, R_k)$.

Thus, the nonlinear system is described by:

$$\begin{cases} \mathbf{x}_k = \mathbf{x}_{k-1} + \Delta t \cdot f(\mathbf{x}_{k-1}, \mathbf{u}_{k-1}) + \mathbf{w}_{k-1} \\ z_k = h(\mathbf{x}_k, \mathbf{u}_k) + \mathbf{v}_k \\ \mathbf{w}_k \sim N(0, Q_k) \\ \mathbf{v}_k \sim N(0, R_k) \end{cases} \quad (17)$$

The initial state is initialized as $\mathbf{x}_0$, and the initial state covariance matrix is set to $P_0$.

The concept of aerodynamic model-aided multi-source data fusion is illustrated in Fig. 4.

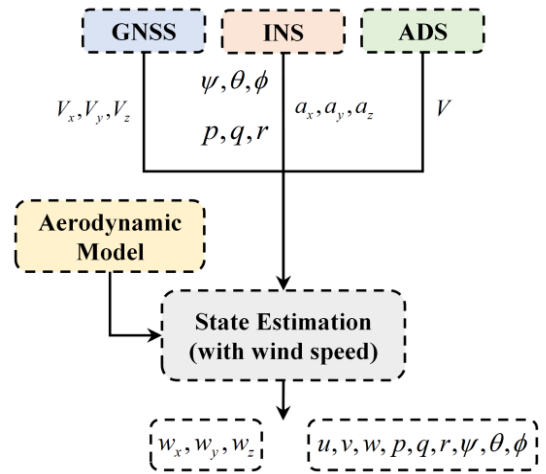

Fig. 4. State and wind estimation using aerodynamic model with low-cost sensors.

The primary steps for state estimation using EKF are:
(1) State Prediction

Given the state estimate at time $t_{k-1}$ as $\hat{x}_{k-1}$, the predicted state at time $t_k$ is obtained from the state prediction equation:

$$\hat{x}_k^- = \hat{x}_{k-1} + \Delta t \cdot f(\hat{x}_{k-1}, u_{k-1}). \tag{18}$$

The covariance matrix is computed as:

$$P_k^- = F_{k-1} P_{k-1} F_{k-1}^T + Q_{k-1} \tag{19}$$

where $F_{k-1}$ denotes the Jacobian matrix of the state prediction equation concerning state variables, i.e.:

$$F_{k-1} = \frac{\partial(\hat{x}_{k-1} + \Delta t \cdot f(\hat{x}_{k-1}, u_{k-1}))}{\partial x}\bigg|_{(\hat{x}_{k-1}, u_{k-1})}. \tag{20}$$

(2) Observation Update

Compute the Kalman gain:

$$K_k = P_k^- H_k^T (H_k P_k^- H_k^T + R_k)^{-1} \tag{21}$$

where $H_k$ denotes the Jacobian matrix of the observation equation concerning state variables, i.e.:

$$H_k = \frac{\partial h}{\partial x}\bigg|_{(\hat{x}_k^-, u_k)}. \tag{22}$$

The state prediction is updated with observations:

$$\hat{x}_k = \hat{x}_k^- + K_k[z_k - h(\hat{x}_k^-, u_k)]. \tag{23}$$

The covariance matrix of the state variables is updated:

$$P_k = (I - K_k H_k) P_k^-. \tag{24}$$

Thus, high-precision state estimation is achieved.

To further enhance data smoothness, a sliding-window moving average estimation is designed. By evaluating the relative relationship between historical data from different phases and a preset threshold, the method augments weights for long-term historical data averages during steady wind conditions to stabilize estimates and reduce high-frequency noise interference. In time-varying wind, it increases weights for short-term historical data averages to track wind variations while mitigating the high-frequency noise introduced by real-time estimates.

Assuming the initial time is $t=0$ and the current time is $t=T_0$ (with the state estimate at time $t=T_0$ denoted as $\hat{x}_k$), two moving windows $D_i = [T_0 - T_i, T_0]$ $(i=1,2)$ are designed, each with a window length $L(D_i) = T_i, T_1 > T_2$. The average state estimate within each window $D_i$ is recorded as $\bar{y}_i$. Due to $T_1 > T_2$, $\bar{y}_1$ represents the long-term historical data average, exhibiting strong stability; whereas $\bar{y}_2$ represents the short-term historical data average, which is closer to the current time $T_0$. This effectively captures recent dynamic variations in the wind and mitigates high-frequency noise compared to the state estimate $\hat{x}_k$ at the current time.

Based on the concept of adaptive estimation, a decision threshold $d$ is set. When the difference between the average historical data $\bar{y}_1$ and $\bar{y}_2$ exceeds $d$, significant dynamic changes in the wind are determined. At this stage, the weight for the state estimate at the most recent time should be increased, i.e., increasing the weight for $\bar{y}_2$. Conversely, when the difference between the average historical data $\bar{y}_1$ and $\bar{y}_2$ does not exceed $d$, the wind is determined to be relatively stable and approximated as a steady wind. Here, the weight for historical state estimates should be increased, i.e., increasing the weight for $\bar{y}_1$. For the state estimate at the current time $t=T_0$, a fixed weight is assigned to stabilize interference from high-frequency noise.

When calculating the smoothed state estimate, different weights $w_i (w_i \geq 0, i=1,2,3)$ are assigned to the referenced historical data $\bar{y}_1$, $\bar{y}_2$, and the current state estimate $\hat{x}_k$. The smoothed state estimate $\hat{x}_k^*$ is computed through a weighted average calculation:

$$\hat{x}_k^* = w_1 \cdot \bar{y}_1 + w_2 \cdot \bar{y}_2 + w_3 \cdot \hat{x}_k \tag{25}$$

$$\begin{aligned} w_1 &= a_1 + \max\left(-a_1, \min\left(a_2, k \cdot \left(d - |\bar{y}_1 - \bar{y}_2|\right)\right)\right) \\ w_2 &= a_2 - \max\left(-a_1, \min\left(a_2, k \cdot \left(d - |\bar{y}_1 - \bar{y}_2|\right)\right)\right) \\ w_3 &= a_3 \end{aligned} \tag{26}$$

where the weights comprise a fixed component and an adaptively adjusted component, $a_1 \sim a_3$ denotes the fixed weight, satisfying: $\sum_{i=1}^{3} a_i = 1 (a_i > 0, i=1,2,3)$. $w_1$ and $w_2$ incorporate the adaptively adjusted component, which represents the product of a scaling coefficient and a normalization factor. From (26), we have: $\sum_{i=1}^{3} w_i = 1 (w_1 \in (0, 1-a_3), w_2 \in (0, 1-a_3), w_3 = a_3 \in (0,1))$.

When $|\bar{y}_1 - \bar{y}_2| > d$, significant dynamic changes in the wind are indicated. At this stage, $k \cdot (d - |\bar{y}_1 - \bar{y}_2|) < 0$ occurs: $w_1$ decreases, $w_2$ increases, making the estimation results more dependent on the short-term historical data average. Conversely, when $|\bar{y}_1 - \bar{y}_2| < d$, the wind is steady. Here, $k \cdot (d - |\bar{y}_1 - \bar{y}_2|) > 0$ occurs: $w_1$ increases, $w_2$ decreases, making the estimation results more dependent on the long-term historical data average.

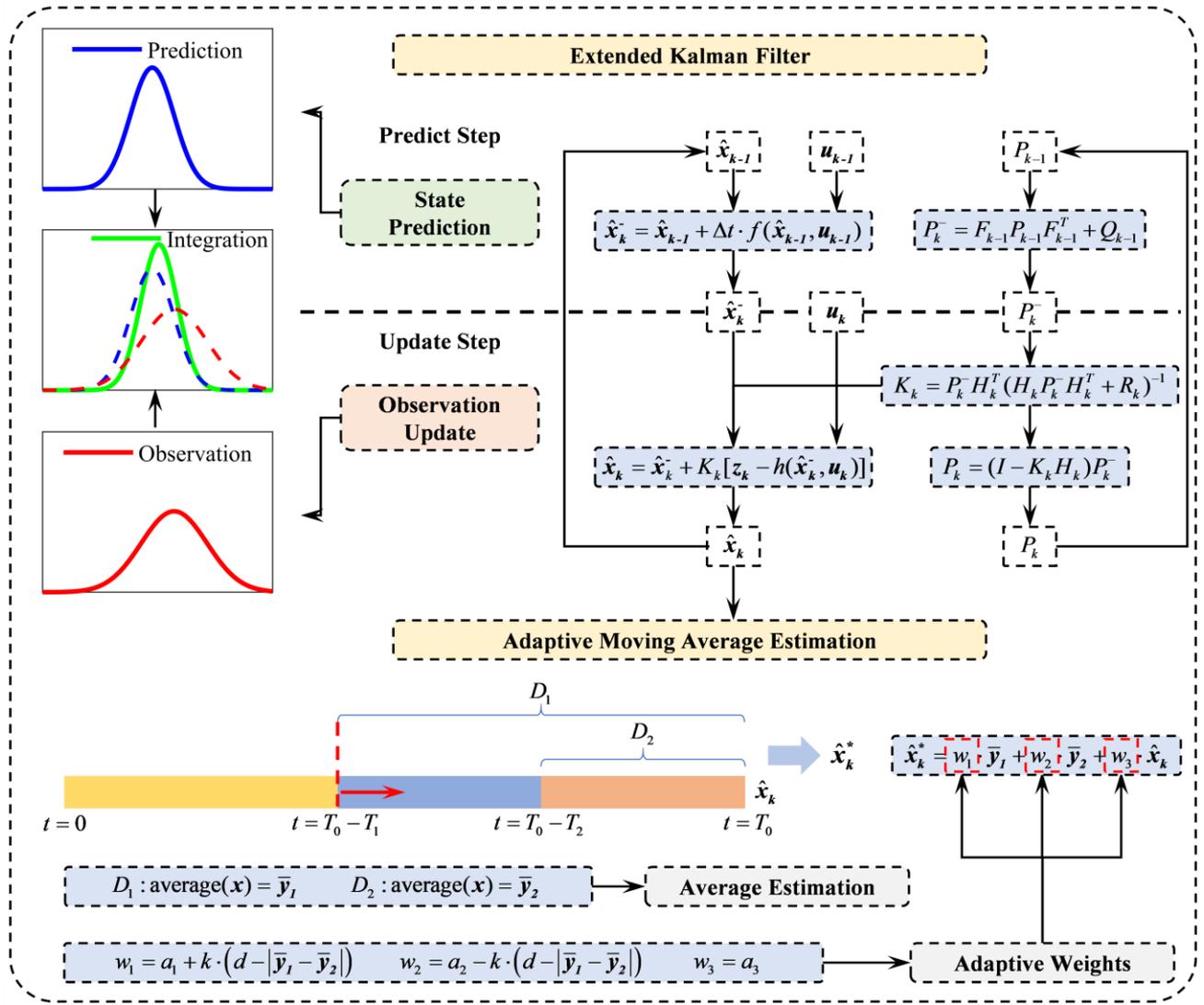

Fig. 5. EKF and AMAE algorithms.

## D. Dryden Wind Model

To construct dynamic wind scenarios that closely resemble realistic atmospheric conditions, this study employs a composite model that combines steady wind with turbulent disturbances. The turbulent component is generated using the Dryden wind turbulence model [34] implemented on the Simulink platform, which utilizes shaping filters to convert Gaussian white noise into turbulence velocity components $(u_g, v_g, w_g)$ with specified power spectral density. Ultimately, the wind vector acting on the UAV is composed of the steady component and turbulent disturbances:

$$\boldsymbol{V}_{wind} = \boldsymbol{V}_{steady} + \left[u_g, v_g, w_g\right]^T. \quad (27)$$

Fig. 6. illustrates the simulated steady wind and disturbed wind, where the average amplitude of wind disturbances is approximately $0.5 \text{ m/s}$, and the maximum disturbance amplitude does not exceed $1 \text{ m/s}$.

## III. SIMULATION EXPERIMENTS

This section conducts simulation studies on UAV wind estimation capabilities based on the dynamic model and state estimation algorithms established in Section II. Considering onboard sensor noise, the effectiveness and noise resistance of the method for filtered estimation of 3D wind in steady wind are first validated. Furthermore, to assess the method's applicability in practical wind conditions, wind estimation simulations are performed under three typical time-varying wind (linearly varying, abruptly changing, and sinusoidally varying wind). The estimation performance is quantitatively evaluated by analyzing root mean square error (RMSE). Finally, aerodynamic model errors are introduced to simulate their impact on wind estimation accuracy, providing insights into potential uncertainties of 3D wind estimation based on small UAVs. All simulations assume the UAV maintains level flight at constant speed at altitude $50\,\text{m}$ with airspeed $16.8\,\text{m/s}$.

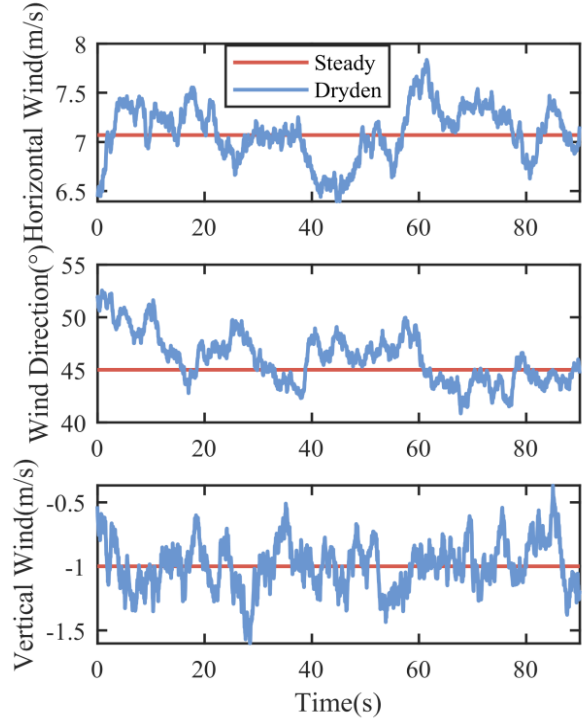

Fig. 6. Comparison between steady and disturbed wind.

### A. Steady Wind Estimation and Analysis

To quantitatively evaluate the filtering performance of the EKF and AMAE algorithms, comparative experiments are conducted within the Simulink simulation framework. The comparison includes:

(1) Direct calculation using sensor data;
(2) EKF estimation;
(3) EKF+AMAE estimation;

Among these, the direct calculation method requires outputting angle of attack and sideslip angle data within the Simulink simulation framework. This enables the computation of wind speed by combining ground speed from GNSS, attitude angles from INS, and airspeed, angle of attack, and sideslip angle measurements from ADS, utilizing the velocity triangle relationship. In contrast, both the EKF estimation method and the EKF+AMAE estimation method operate under conditions where flow angle sensors in the ADS are failed or absent—that is, without utilizing angle of attack or sideslip angle data. The estimation results are compared with the calculation results to assess the accuracy and errors of the algorithms.

The steady wind configuration for simulations is specified in Table II: Wind velocity in $x, y$ directions are both $5\,\text{m/s}$, equivalent to Beaufort scale level 4; while in $z$ direction the upwash intensity is $1\,\text{m/s}$.

**TABLE II**
**Steady Wind**

| Wind component | Value |
|---|---|
| $w_x$ | 5m/s |
| $w_y$ | 5m/s |
| $w_z$ | −1m/s |

Fig. 7. illustrates the estimation performance for a steady wind. Subfigures (a), (c), and (e) present the estimation results for horizontal wind speed, horizontal wind direction, and vertical wind speed, respectively. Among these subfigures, the black curve represents the true wind, the gray curve represents the wind directly calculated using sensor data, the blue curve represents the wind estimated using the EKF method, and the red curve represents the wind estimated using the EKF method and subsequently smoothed using the AMAE approach. Subfigures (b), (d), and (f) depict the estimation errors for horizontal wind speed, horizontal wind direction, and vertical wind speed, respectively. These errors represent the difference between the wind estimated by different methods and the true values. The black curve in these subfigures represents the reference value of 0 (indicating that, ideally, the estimation error should be zero when the estimated result matches the true result).

Subfigure (a) indicates that the calculated and estimated horizontal wind speed is approximately $7\,\text{m/s}$, consistent with the characteristic of a steady wind perturbed by the Dryden wind model. Subfigure (c) shows that the estimated horizontal wind direction generally aligns with the actual wind direction. Subfigure (e) indicates that the calculated and estimated upwash flow strength is approximately $1\,\text{m/s}$. Analysis of the error levels calculated from subfigures (b), (d), and (f) reveals that the overall wind estimation error using the direct calculation method is relatively high. Employing the EKF method and the EKF+AMAE method significantly reduces the wind estimation error level, even without an angle-of-attack sensor. Furthermore, for horizontal wind speed estimation, the EKF+AMAE method achieves a specific reduction in error compared to the EKF method alone. Regarding the estimation of horizontal wind direction and vertical wind speed, the error levels between the EKF and EKF+AMAE methods are comparable.

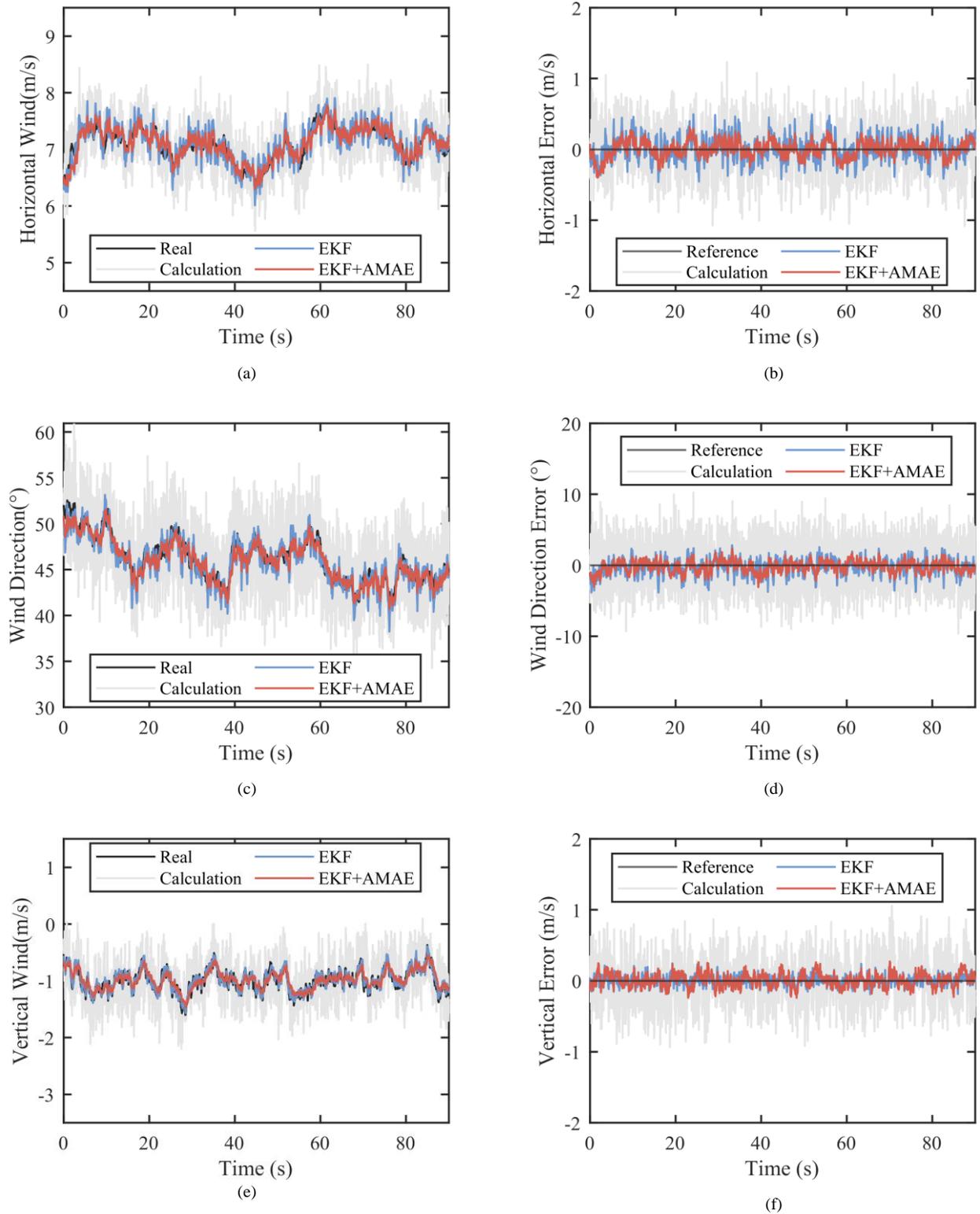

Fig. 7. Steady wind estimation. (a) Estimation of horizontal wind speed. (b) Error of horizontal wind speed. (c) Estimation of horizontal wind direction. (d) Error of horizontal wind direction. (e) Estimation of vertical wind speed. (f) Error of vertical wind speed.

The RMSE results are presented in Table III. The direct calculation yields an RMSE of about 0.3 m/s for both horizontal and vertical wind speed, and an RMSE of about 3° for horizontal wind direction. The RMSE for horizontal wind speed estimated using the EKF method is reduced to 0.1682m/s. Employing the EKF+AMAE method further minimizes the RMSE for horizontal wind speed estimation to $0.1150 \, \text{m/s}$. Both estimation methods achieve an RMSE of approximately $1°$ for horizontal wind direction estimation and an RMSE of less than $0.1 \, \text{m/s}$ for vertical speed estimation.

**TABLE III**
**RMSE of Steady Wind Estimation**

| Wind component | Calculation | EKF | EKF+AMAE |
|---|---|---|---|
| Horizontal wind $(\text{m/s})$ | 0.3023 | 0.1682 | 0.1150 |
| Horizontal wind $(°)$ | 2.5562 | 1.1121 | 0.8339 |
| Vertical wind $(\text{m/s})$ | 0.2939 | 0.0685 | 0.0906 |

This demonstrates that due to the presence of sensor noise, the direct calculation of wind speed using GNSS, INS, and ADS sensor data yields significant errors. In the absence of the angle-of-attack and sideslip angle sensors within the ADS, the direct calculation method is particularly unsuitable for wind estimation. Conversely, employing state estimation methods such as EKF and EKF+AMAE enables wind estimation even when the angle-of-attack and sideslip angle sensors in the ADS fail or are absent. The filtering effectiveness is pronounced, significantly mitigating the impact of sensor noise and reducing the fluctuating characteristics of the estimation results, thereby demonstrating excellent filtering performance. Furthermore, the introduction of the AMAE smoothing method enhances the accuracy of the estimation results compared to using the EKF method alone, particularly when facing larger reference wind speeds.

**B. Time-Varying Wind Estimation**

The following research investigates the estimation performance of this wind estimation method in three typical time-varying wind environments (including linear wind, abruptly changing wind, and sinusoidal wind). Due to the estimation characteristics of the technique, the scenario where a UAV experiences time-varying wind during constant-altitude and constant-speed flight can be conceptually likened to the scenario where the UAV experiences spatially varying but time-invariant wind when traversing different spatial regions. Therefore, focusing solely on time-varying wind estimation allows the results to be generalized to wind estimation in spatially heterogeneous conditions. Since the error characteristics of the direct calculation method were analyzed in the previous section, concluding that both the EKF and EKF+AMAE methods exhibit superior filtering performance, the direct calculation results will not be plotted in the following analysis for the sake of clarity and conciseness in the figures.

First, the linear wind is investigated. A linear wind denotes a scenario in which the wind speed increases or decreases linearly over a specified period. This can be conceptually likened to the UAV traversing spatial regions characterized by linearly differing wind speeds. When the UAV performs climbing or descending maneuvers, the spatial variation in wind speed manifests as differences along the vertical direction, known as a wind gradient [35]. Wind gradients are typical in dynamic soaring flight scenarios. Furthermore, the linear wind gradient is a commonly adopted assumption in research on dynamic soaring. Therefore, estimating linear wind is essential.

**TABLE IV**
**Linear Wind**

| Wind component | Value |
|---|---|
| $w_x$ | $2 \sim 8 \, \text{m/s}$ |
| $w_y$ | $2 \sim 5 \, \text{m/s}$ |
| $w_z$ | $-1 \sim -2.5 \, \text{m/s}$ |

The parameters of the linear wind adopted in this study are detailed in Table IV. During $t \leq 30\text{s}$ and $60\text{s} \leq t \leq 90\text{s}$, the environment maintains a steady wind. The wind speed changes in all three directions during $30\text{s} \leq t \leq 60\text{s}$ are described by (28):

$$w_x = 2 + 0.2 \cdot (t - 30)$$
$$w_y = 2 + 0.1 \cdot (t - 30) \quad (28)$$
$$w_z = -1 - 0.05 \cdot (t - 30).$$

Fig. 8. presents the estimation results for the linear wind using the EKF and EKF+AMAE methods. Subfigures (a) and (e) demonstrate the transition of the estimated wind characteristics from a steady wind to a linear wind and back to a steady wind. Subfigure (c) shows the horizontal wind direction. Within the initial $30\text{s}$, a persistent change in wind direction is observed. This occurs because, under low reference wind speeds, the disturbances introduced by the Dryden wind model become significant. This leads to instability in the resultant horizontal wind direction after wind speed changes in $x, y$ directions and concurrently results in larger estimation errors for wind direction. This is reflected in subfigure (d), where the error level during the initial $30\text{s}$ is higher compared to that within $30 \sim 90\text{s}$. The error levels calculated from subfigures (b) and (f) indicate that the wind estimation errors using both the EKF method and the EKF+AMAE method remain at a low level.

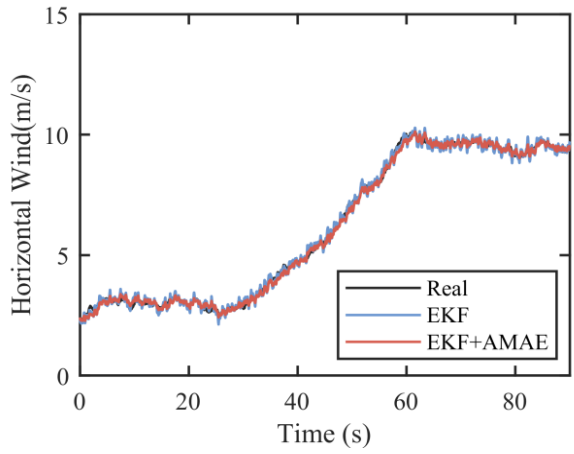
(a)

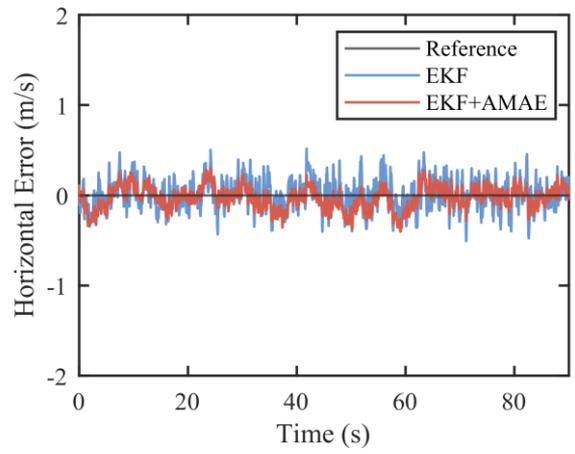
(b)

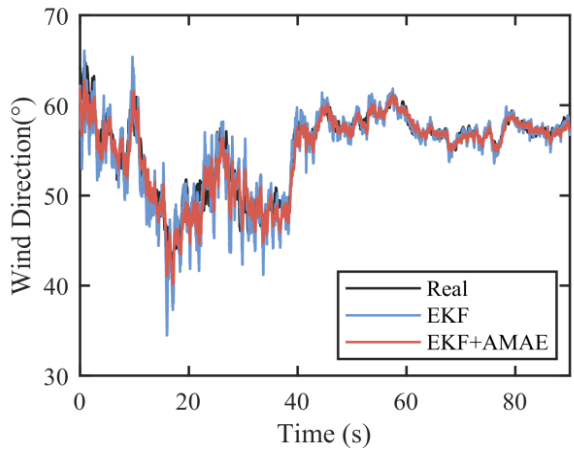
(c)

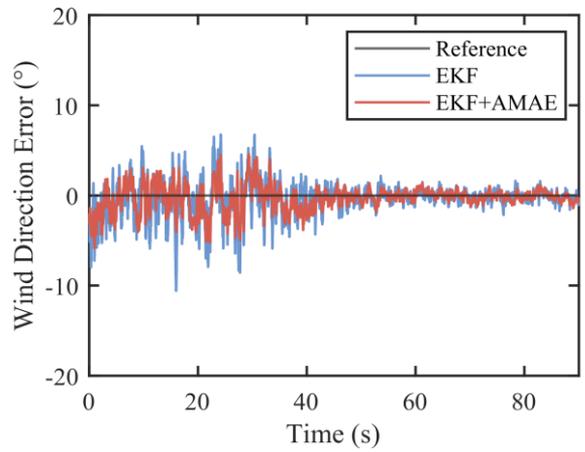
(d)

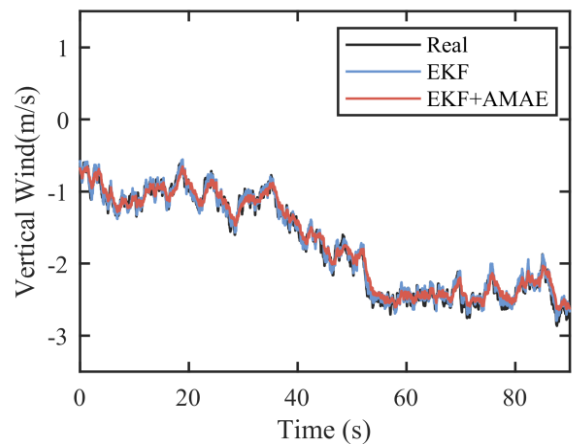
(e)

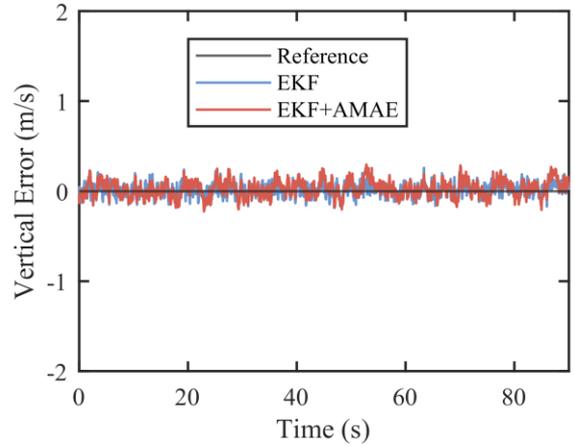
(f)

Fig. 8. Linear wind estimation. (a) Estimation of horizontal wind speed. (b) Error of horizontal wind speed. (c) Estimation of horizontal wind direction. (d) Error of horizontal wind direction. (e) Estimation of vertical wind speed. (f) Error of vertical wind speed.

The RMSE calculation results for the linear wind estimation are presented in Table V. Both methods yield an RMSE of less than 0.2 m/s for horizontal wind speed estimation, an RMSE within $2°$ for horizontal wind direction estimation, and an RMSE below 0.1 m/s for vertical wind speed estimation.

**TABLE V**
**RMSE of Linear Wind Estimation**

| Wind component | EKF | EKF+AMAE |
|---|---|---|
| Horizontal wind (m/s) | 0.1662 | 0.1275 |
| Horizontal wind (°) | 1.9350 | 1.4293 |
| Vertical wind (m/s) | 0.0701 | 0.0914 |

The above estimation results and RMSE values demonstrate that the algorithm provides stable and accurate estimation performance in time-varying wind, effectively tracking changes in wind speed promptly.

The following section investigates abruptly changing wind, characterized by abrupt increases or decreases in wind speed over short time intervals. This phenomenon is analogous to wind speed variations encountered by UAVs within minimal altitude changes, a feature known as wind shear [36]. Wind shear is prevalent in mountainous terrains and represents a typical wind condition exploitable for UAV dynamic soaring. Therefore, precise estimation of such wind is essential.

**TABLE VI**
**Abruptly Changing Wind**

| Wind component | Value |
|---|---|
| $w_x$ | 2 ~ 7 m/s |
| $w_y$ | 2 ~ 7 m/s |
| $w_z$ | −1 ~ −2 m/s |

The parameters of the abruptly changing wind adopted in this study are detailed in Table VI. During $t \leq 30s$ and $60s \leq t \leq 90s$, the environment maintains a steady wind. In $30s \leq t \leq 60s$, wind speed variations along all three axes follow (29), which ensures a smooth transition between steady states through exponential delay. Wind speed in $x, y$ directions undergo a transition from steady component $2$ m/s to $7$ m/s within 10s and remains steady thereafter. Similarly, wind speed in $z$ direction transitions from steady component $-1$ m/s to $-2$ m/s within 10s and maintains steady conditions abruptly.

$$w_x = 2 + \frac{5}{1+e^{-0.7 \cdot (t-35)}}$$
$$w_y = 2 + \frac{5}{1+e^{-0.7 \cdot (t-35)}} \quad (29)$$
$$w_z = -1 - \frac{1}{1+e^{-0.7 \cdot (t-35)}}.$$

Fig. 9. presents the estimation results for the abruptly changing wind using the EKF and EKF+AMAE methods. Subfigures (a) and (e) demonstrate the abrupt transition of the estimated wind characteristics from a smaller steady wind to a larger steady wind within 10s. Subfigure (c) shows the horizontal wind direction. The directional variation observed in the initial 30s shares the same underlying cause as during the linear wind phase and similarly results in larger wind direction estimation errors. As shown in subfigures (b) and (f), the wind estimation errors using both the EKF and EKF+AMAE methods remain at a low level.

The RMSE calculation results for the abruptly changing wind estimation are presented in Table VII. Both methods achieve an RMSE of less than $0.2$ m/s for horizontal wind speed estimation, an RMSE within $2°$ for horizontal wind direction estimation, and an RMSE below $0.1$ m/s for vertical wind speed estimation. These results indicate that the wind estimation method is effectively validated in abruptly changing wind.

**TABLE VII**
**RMSE of Abruptly Changing Wind Estimation**

| Wind component | EKF | EKF+AMAE |
|---|---|---|
| Horizontal wind (m/s) | 0.1745 | 0.1294 |
| Horizontal wind (°) | 1.4417 | 0.8809 |
| Vertical wind (m/s) | 0.0691 | 0.0920 |

The sinusoidal wind is now investigated. By estimating sinusoidal wind, the scope of wind estimation can be extended to arbitrarily varying wind, demonstrating broad application prospects. Table VIII details the sinusoidal wind conditions: $x, y$ directions wind speed vary between $3 \sim 7$ m/s, while $z$ direction wind speed varies between $-1.5 \sim -0.5$ m/s. The sinusoidal wind configuration is given by (30).

$$w_x = 5 + 2\sin(0.1t+\varphi)$$
$$w_y = 5 + 2\cos(0.1t+\varphi) \quad (30)$$
$$w_z = -1 + 0.5\sin(0.1t+\varphi).$$

**TABLE VIII**
**Sinusoidal Wind**

| Wind component | Value |
|---|---|
| $w_x$ | 3 ~ 7 m/s |
| $w_y$ | 3 ~ 7 m/s |
| $w_z$ | −1.5 ~ −0.5 m/s |

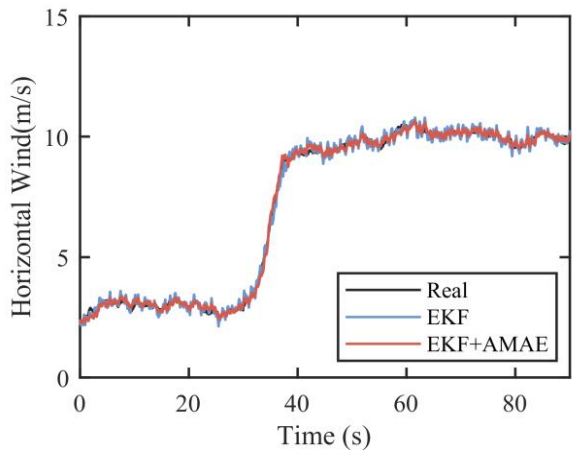
(a)
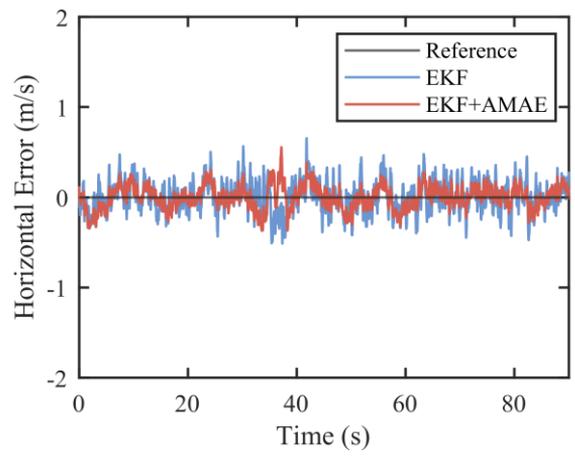
(b)
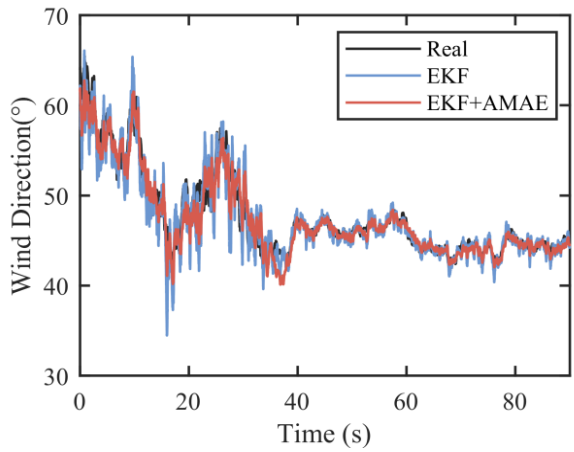
(c)
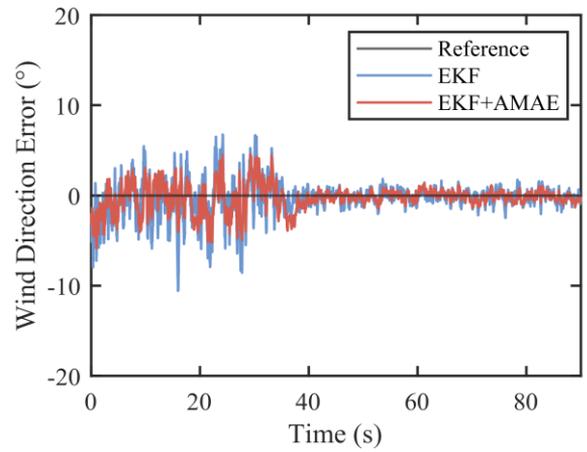
(d)
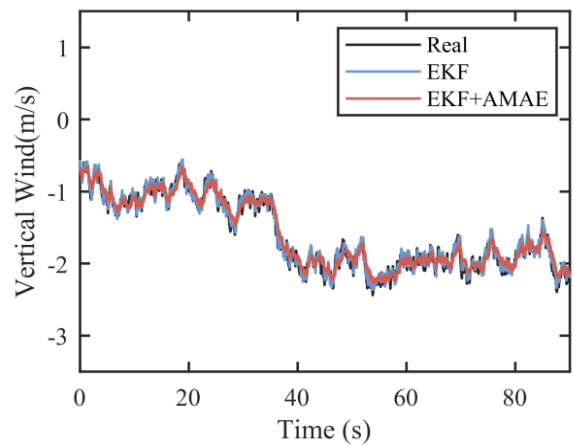
(e)
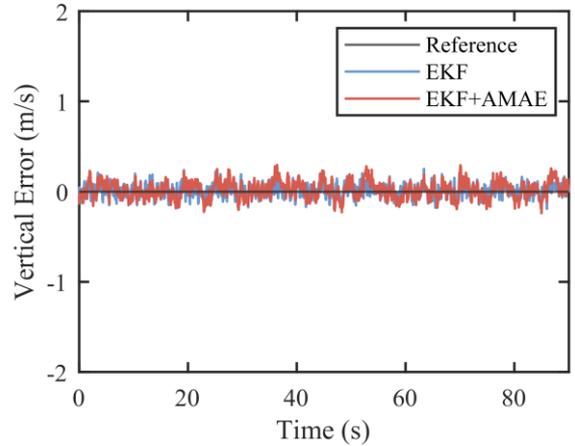
(f)

Fig. 9. Abruptly changing wind estimation. (a) Estimation of horizontal wind speed. (b) Error of horizontal wind speed. (c) Estimation of horizontal wind direction. (d) Error of horizontal wind direction. (e) Estimation of vertical wind speed. (f) Error of vertical wind speed.

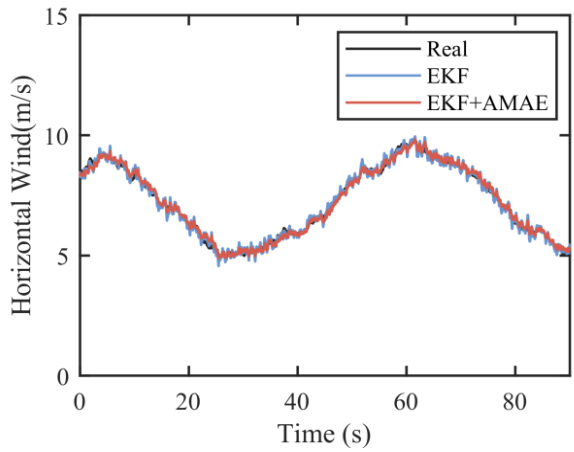
(a)

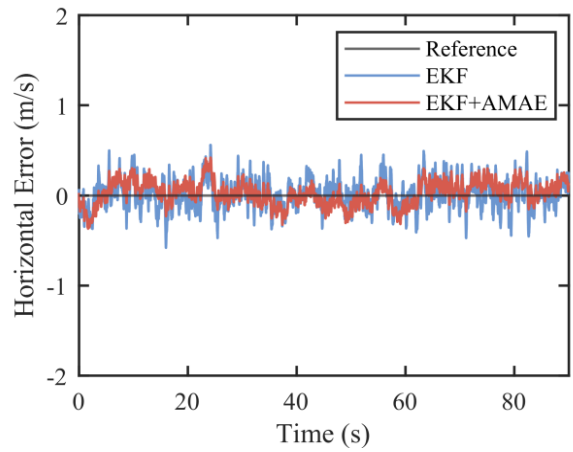
(b)

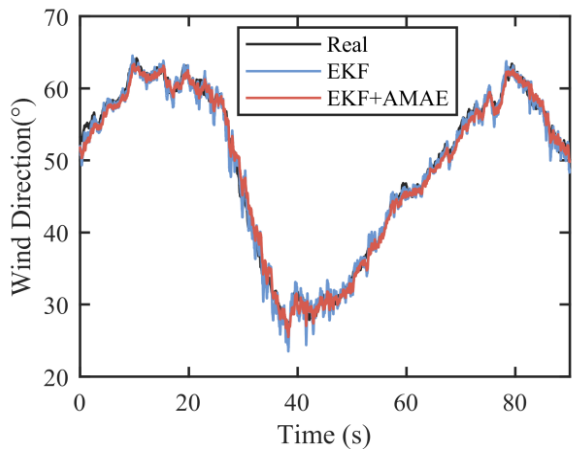
(c)

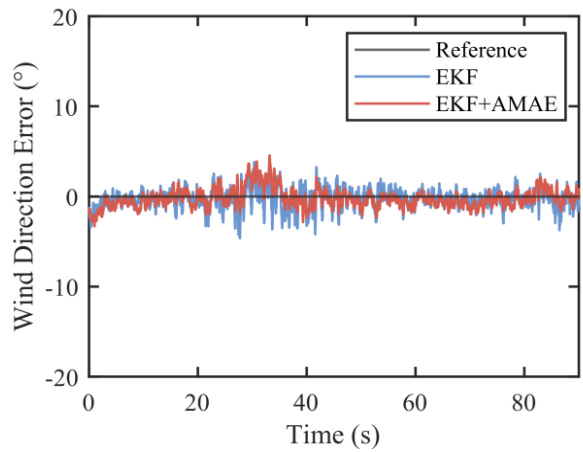
(d)

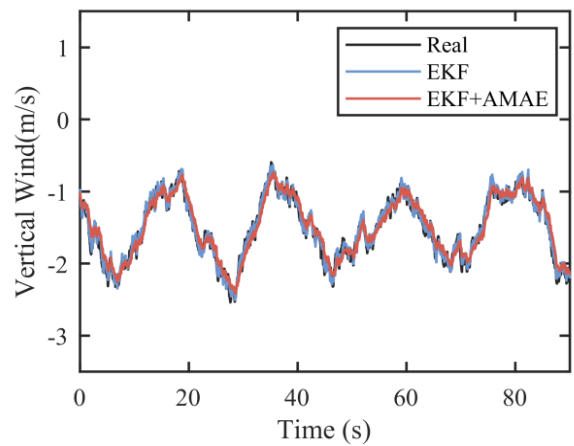
(e)

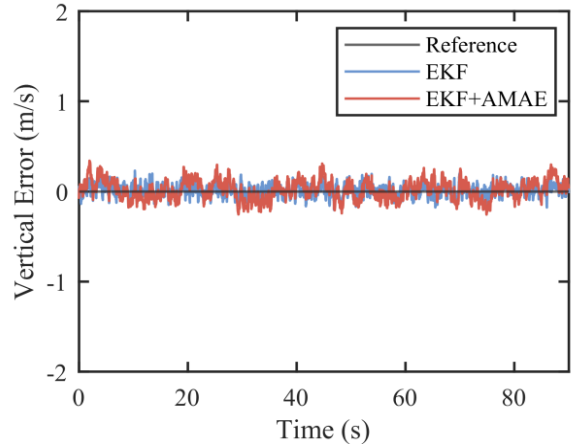
(f)

Fig. 10. Sinusoidal wind estimation. (a) Estimation of horizontal wind speed. (b) Error of horizontal wind speed. (c) Estimation of horizontal wind direction. (d) Error of horizontal wind direction. (e) Estimation of vertical wind speed. (f) Error of vertical wind speed.

Fig. 10. presents the estimation results for the sinusoidal wind using the EKF and EKF+AMAE methods. Subfigures (a) and (e) demonstrate that the estimated wind exhibit distinct sinusoidal variation characteristics. Subfigure (c) shows the directional variation of the horizontal wind. Subfigures (b), (d), and (f) display the error curves for wind speed and direction estimation.

**TABLE IX**
**RMSE of Sinusoidal Wind Estimation**

| Wind component | EKF | EKF+AMAE |
|---|---|---|
| Horizontal wind $(m/s)$ | 0.1700 | 0.1352 |
| Horizontal wind $(°)$ | 1.1599 | 1.0546 |
| Vertical wind $(m/s)$ | 0.0700 | 0.1065 |

The RMSE calculation results for the sinusoidal wind estimation are provided in Table IX. Both methods achieve an RMSE of less than $0.2$ m/s for horizontal wind speed estimation, an RMSE within $2°$ for horizontal wind direction estimation, and an RMSE of about $0.1$ m/s for vertical wind speed estimation.

Based on the above results, the algorithm demonstrates favorable estimation performance for sinusoidal wind. It is therefore reasonable to infer that the algorithm can achieve accurate estimation of arbitrarily varying wind with an estimated accuracy of $0.2$ m/s.

Integrating the preceding analysis and research, this section performs wind estimation for typical time-varying wind using the EKF and EKF+AMAE methods under missing angle-of-attack and sideslip angle data in the ADS. The estimation results are accurate, with RMSEs for both horizontal and vertical wind speed estimation not exceeding $0.2$ m/s, RMSE for horizontal wind direction estimation not exceeding $2°$, and estimation latency at the second level. Furthermore, compared to the EKF method alone, the EKF+AMAE method achieves lower RMSE values and higher estimation accuracy, validating the effectiveness of the wind estimation approach.

### C. Impact of Aerodynamic Model Accuracy on Wind Estimation Errors

**TABLE X**
**Relationship Between $C_{L0}$ Error and Wind Estimation Errors**

| $C_{L0}$ error | Estimation results | | RMSE | | |
|---|---|---|---|---|---|
| | $\alpha(°)$ | $C_L$ | Horizontal wind (m/s) | Horizontal direction $(°)$ | Vertical wind (m/s) |
| 0.15 | 1.4250 | 0.3032 | 0.1389 | 1.4304 | 0.3890 |
| 0.12 | 1.9089 | 0.3020 | 0.1321 | 1.2792 | 0.3186 |
| 0.09 | 2.3925 | 0.3008 | 0.1263 | 1.1384 | 0.2488 |
| 0.06 | 2.8759 | 0.2996 | 0.1214 | 1.0127 | 0.1813 |
| 0.03 | 3.3558 | 0.2984 | 0.1176 | 0.9091 | 0.1222 |
| 0.00 | 3.8419 | 0.2971 | 0.1150 | 0.8339 | 0.0906 |
| -0.03 | 4.3214 | 0.2959 | 0.1136 | 0.7975 | 0.1153 |
| -0.06 | 4.8072 | 0.2946 | 0.1133 | 0.8041 | 0.1754 |
| -0.09 | 5.2895 | 0.2934 | 0.1142 | 0.8529 | 0.2467 |
| -0.12 | 5.7716 | 0.2921 | 0.1162 | 0.9371 | 0.3227 |
| -0.15 | 6.2535 | 0.2908 | 0.1192 | 1.0479 | 0.4012 |

Simulation studies reveal that the accuracy of the aerodynamic model impacts wind estimation results. However, for small UAVs, obtaining precise aerodynamic models (e.g., through wind tunnel testing) incurs significant costs. Furthermore, simplifications made during the modeling process lead to discrepancies between the aerodynamic forces estimated by the model and those generated in actual flight environments. Such model inaccuracies affect state estimation values during prediction. To gain clearer insight into the error levels of estimated wind, this section focuses on analyzing the impact of aerodynamic model errors on wind estimation. The wind in the simulation environment is configured as a steady wind, with parameters detailed in Table II.

First, the influence of the initial $C_{L0}$ error on wind estimation is investigated. Table X presents the effect of $C_{L0}$ error on wind estimation errors. It should be noted that during flight simulation, the measured angle of attack output is $\alpha = 3.8653°$, and the calculated lift coefficient is $C_L = 0.2964$. These values can be compared with the estimated angle of attack and lift coefficient values in Table X.

As can be seen from Table X, as the error in $C_{L0}$ gradually increases or decreases from zero, the estimate of $C_L$ remains centered around the true value, while the estimated $\alpha$ exhibits a significant deviation from the true value. Regarding wind estimation, the speed and direction of the horizontal wind maintain a relatively stable level of error. Specifically, the RMSE of the horizontal wind speed remains within $0.2 \text{ m/s}$ with minor error variations, and the RMSE of the horizontal wind direction does not exceed $2°$. However, the RMSE of the vertical wind speed error shows substantial deviation following changes in the angle of attack estimate. When the initial error in $C_{L0}$ is $\pm 0.15$, the RMSE of the estimated vertical wind speed already reaches $0.4 \text{ m/s}$. The impact of the error in $C_{L0}$ on the estimation of flight state variables and the wind estimation error is presented as bar charts and line charts, respectively, as shown in Fig. 11.

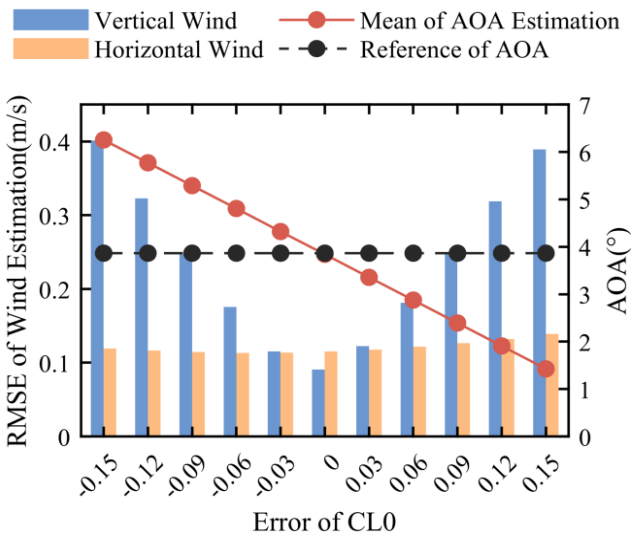

Fig. 11. Impact of $C_{L0}$ error on state and wind estimation.

In Fig. 11., the blue bars represent the RMSE of the estimated vertical wind speed under different $C_{L0}$ errors, while the orange bars represent the RMSE of the estimated horizontal wind speed under different $C_{L0}$ errors. It can be observed that the RMSE of the vertical wind estimation varies drastically, exhibiting larger estimation errors when $C_{L0}$ error is significant. In contrast, the horizontal wind estimation error remains stable and is largely unaffected by $C_{L0}$ error. The black curve denotes the reference angle of attack, i.e., the measured angle of attack output during the simulation. The red curve represents the average value of the estimated angle of attack. It can be seen that as $C_{L0}$ error increases, the error in the estimated angle of attack also gradually increases.

Fundamentally, this error arises from the correction mechanism of the accelerometer. When an $C_{L0}$ error exists, assuming the estimated value of $C_{L0}$ is too large, this leads to an overestimation of $C_L$, consequently causing an overestimation of lift. However, the accelerometer corrects the lift, reducing the estimation of $\alpha$ a key factor in maintaining accurate lift estimation. Under the constraint of airspeed, reducing the estimation of $\alpha$ causes $u$ to increase and $w$ to decrease, with the change in $w$ being significantly greater than that in $u$, while $v$ remains essentially constant. Consequently, when the estimated value of $C_{L0}$ is too large, it primarily results in an underestimation of the airspeed component $w$. Combined with the velocity triangle relationship, this leads to an overestimation of the vertical wind component $w_z$. Since the estimated value of $u$、$v$ shows negligible change, the error introduced by the lift model has almost no impact on the estimation of the speed and direction of the horizontal wind. A similar analysis applies when the estimated value of $C_{L0}$ is too small.

Further, the impact of $C_{L\alpha}$ error on wind estimation can be investigated. Table XI presents the influence of the error level in $C_{L\alpha}$ on wind estimation error.

The variation characteristics of the estimated values shown in Table XI are mainly consistent with those in Table X. As $C_{L\alpha}$ error gradually increases from 0, the estimate of $C_L$ remains centered around the true value, whereas the estimated angle of attack exhibits a significant deviation from the true value. Regarding wind estimation, the errors are primarily manifested in the vertical direction. The impact of $C_{L\alpha}$ error on the estimation of flight state variables and the wind estimation error is presented as bar charts and line charts, as shown in Fig. 12. In Fig. 12., the RMSE of the estimated vertical wind speed (represented by the blue bars) and the estimated angle of attack (represented by the red curve) both show significant variations with $C_{L\alpha}$ error. In contrast, the horizontal wind estimation error (represented by the orange bars) remains relatively stable and is largely unaffected by $C_{L\alpha}$ error.

**TABLE XI**
**Relationship Between $C_{L\alpha}$ Error and Wind Estimation Errors**

| $C_{L\alpha}$ error | Estimation results | | RMSE | | |
| --- | --- | --- | --- | --- | --- |
| | $\alpha\,(°)$ | $C_L$ | Horizontal wind (m/s) | Horizontal direction (°) | Vertical wind (m/s) |
| 4 | 1.8083 | 0.3021 | 0.1325 | 1.3307 | 0.3174 |
| 3 | 2.0817 | 0.3015 | 0.1292 | 1.2395 | 0.2809 |
| 2 | 2.4553 | 0.3006 | 0.1251 | 1.1273 | 0.2313 |
| 1 | 2.9954 | 0.2993 | 0.1202 | 0.9890 | 0.1619 |
| 0 | 3.8419 | 0.2971 | 0.1150 | 0.8339 | 0.0906 |
| -1 | 5.8042 | 0.2911 | 0.1218 | 1.0980 | 0.3271 |

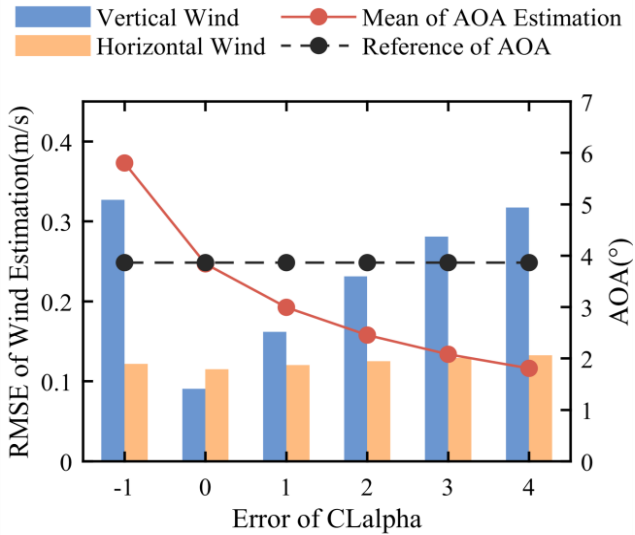

Fig. 12. Impact of $C_{L\alpha}$ error on state and wind estimation.

The error caused by inaccuracies in $C_{L\alpha}$ is similar in nature to that caused by inaccuracies in $C_{L0}$. When the estimated value of $C_{L\alpha}$ is too large, a reduction in the estimation of $\alpha$ is required to ensure accurate lift estimation. Consequently, this results in an overestimation of $w_z$. Due to the multiplicative relationship between $C_{L\alpha}$ and $\alpha$, and given $C_{L\alpha} > 0$, a decrease in $C_{L\alpha}$ leads to a significant increase in the estimated value of $\alpha$. Consequently, the error in the vertical wind estimation also increases in magnitude. Therefore, the relationship between the vertical wind estimation error and $C_{L\alpha}$ error around 0 is not symmetric and linear; instead, it essentially exhibits an inverse relationship.

In addition to lift model errors, the impact of drag model errors on wind estimation can also be investigated. Table XII presents the influence of the error level in $C_{D0}$ on wind estimation error. It should be additionally noted that the drag coefficient value calculated during the flight simulation is $C_D = 0.0331$, which can be used for comparison with the estimated drag coefficient value.

As can be seen from Table XII, as $C_{D0}$ error gradually increases or decreases from 0, the estimated value of $C_D$ deviates from the true value accordingly. In contrast, both the estimated $\alpha$ and $C_L$ closely match their true values. Regarding wind estimation, the error in the estimated speed of the horizontal wind exhibits some fluctuation but remains within 0.2 m/s. The error in the estimated direction of the horizontal wind is around $2°$, maintaining a relatively small range. However, the RMSE of the vertical wind speed error shows significant deviation corresponding to the error in the zero-lift drag coefficient. When $C_{D0}$ error reaches $\pm 0.01$, the RMSE of the estimated vertical wind speed has already exceeded 0.5 m/s.

The impact of $C_{D0}$ error on the estimation of flight state variables and on the wind estimation error is presented as bar charts and line charts, as shown in Fig. 13.

In Fig. 13., the blue bars represent the RMSE of the estimated vertical wind speed under different $C_{D0}$ errors, while the orange bars represent the RMSE of the estimated horizontal wind speed under different $C_{D0}$ errors. It can be observed that the RMSE of the vertical wind speed estimation varies drastically, exhibiting larger estimation errors when $C_{D0}$ error is significant. In contrast, the horizontal wind speed estimation error does not fluctuate as severely as its vertical counterpart; however, compared to the horizontal wind speed estimation error induced by lift model inaccuracies, it introduces noticeable errors. The black curve denotes the reference drag coefficient, i.e., the calculated drag coefficient output during the simulation. The red curve represents the average value of the estimated drag coefficient. It can be seen that as $C_{D0}$ error increases, the error in the estimated drag coefficient also gradually increases.

TABLE XII
Relationship Between $C_{D0}$ Error and Wind Estimation Errors

| $C_{D0}$ error | Estimation results | RMSE | | |
|---|---|---|---|---|
| | $C_D$ | Horizontal wind (m/s) | Horizontal direction (°) | Vertical wind (m/s) |
| 0.010 | 0.0433 | 0.1663 | 2.2157 | 0.5050 |
| 0.008 | 0.0415 | 0.1494 | 1.8789 | 0.4204 |
| 0.006 | 0.0392 | 0.1331 | 1.5022 | 0.3156 |
| 0.004 | 0.0371 | 0.1228 | 1.2055 | 0.2208 |
| 0.002 | 0.0352 | 0.1170 | 0.9827 | 0.1373 |
| 0.000 | 0.0330 | 0.1150 | 0.8339 | 0.0906 |
| -0.002 | 0.0309 | 0.1165 | 0.8098 | 0.1459 |
| -0.004 | 0.0289 | 0.1205 | 0.8853 | 0.2415 |
| -0.006 | 0.0268 | 0.1258 | 1.0176 | 0.3472 |
| -0.008 | 0.0250 | 0.1313 | 1.1546 | 0.4459 |
| -0.010 | 0.0227 | 0.1380 | 1.3206 | 0.5694 |

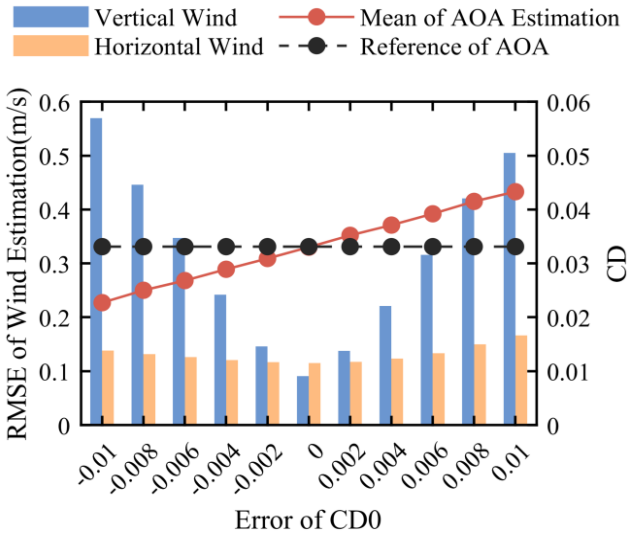

Fig. 13. Impact of $C_{D0}$ error on state and wind estimation.

Unlike the error caused by lift model inaccuracies, when an error exists in $C_{D0}$, assuming the estimated value of $C_{D0}$ is too large, the estimated value of $C_D$ will also be too large. At this point, due to $C_L$ estimation, $\alpha$ cannot be adjusted downward to correct the drag estimation. Consequently, the overestimated drag calculation leads to an underestimation of $u$. Under the constraint of $\alpha$, the estimated value of $w$ becomes too small, while $w_z$ becomes too large. Additionally, due to the constraint of airspeed, when the estimated value of $u, w$ are too small, the estimated value of $v$ becomes too large, resulting in errors in the horizontal wind estimation. A similar analysis applies when the estimated value of $C_{D0}$ is too small.

In summary, this section analyzed the underlying mechanisms by which inaccuracies in the aerodynamic model, specifically lift model errors and drag model errors, lead to wind estimation errors. This analysis provides a basis for evaluating the application potential of this wind estimation method on small UAV platforms.

## IV. FLIGHT TEST

### A. Test Conditions

To validate the practical application performance of this wind estimation method, this study further conducted flight tests for wind estimation. The Skywalker X8 was employed as the carrier platform. This UAV features a tailless flying-wing configuration, offering a high lift-to-drag ratio, excellent flight stability, easy maneuverability, and high safety. Its modular payload compartment design supports the rapid integration and layout optimization of various sensor types, providing a hardware foundation for the flexible deployment of wind detection equipment. The UAV platform and onboard equipment configuration are shown in Fig. 14.:

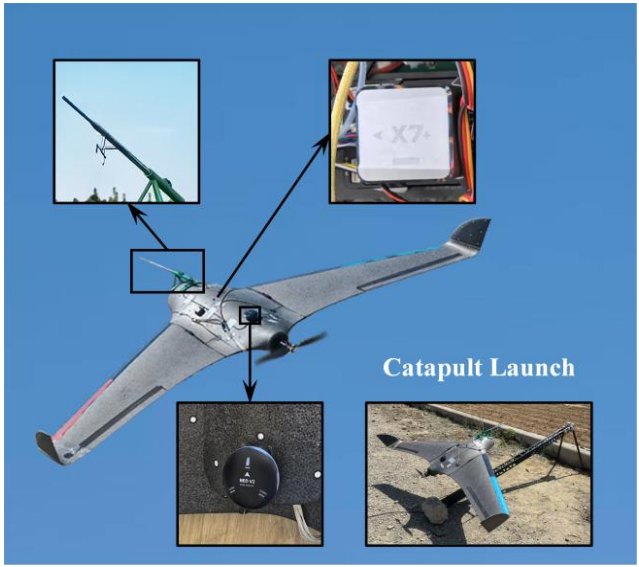

Fig. 14. Skywalker X8 and onboard equipment.

Fig. 14. specifically highlights the ADS, flight controller, and GPS mounted on the UAV. The ADS incorporates a vane anemometer for directly calculating wind speed, which is compared with the estimation results, operating at a sampling frequency of 100 Hz. The flight controller utilizes the open-source CUAV X7+, also sampling at 100 Hz. The GPS unit is an NEO V2, sampling at 5 Hz; data alignment during processing was achieved via linear interpolation.

Due to differences in the onboard equipment added, the actual measured weight of the entire aircraft was $4.48 \text{ kg}$. Other key parameters remained consistent with those listed in Table I.

Existing research on the aerodynamic model for the Skywalker X8 is also very comprehensive and mature. This study utilized available wind tunnel test data [37] as the aerodynamic model for wind estimation during the flight tests. Key aerodynamic force and moment coefficients are presented in Table XIII.

### B. Flight Test Results and Analysis

During the flight tests, the environment maintained a predominantly westerly horizontal wind of around 10 m/s and an updraft of around 2 m/s, accompanied by unstable gusts. The UAV maintained level flight at constant speed and altitude 270 m from northeast to southwest at an airspeed of 27 m/s, collecting test data which was subsequently processed offline. A mobile device equipped with an ARM platform was used for data processing. Processing the flight test data for 27s took 1.15s, demonstrating the potential for real-time onboard computation with this method.

**TABLE XIII**
**Coefficients of Skywalker X8**

| Coefficient | Value | Coefficient | Value |
|---|---|---|---|
| $C_{L0}$ | 0.0867 | $C_{C\delta_a}$ | 0.0433 |
| $C_{L\alpha}$ | 4.02 | $C_{Cp}$ | - |
| $C_{L\delta_e}$ | 0.278 | $C_{Cr}$ | - |
| $C_{Lq}$ | - | $C_{l0}$ | 0.00413 |
| $C_{D0}$ | 0.0197 | $C_{l\beta}$ | -0.0849 |
| $C_{D\alpha}$ | 0.0791 | $C_{l\delta_a}$ | 0.12 |
| $C_{D\delta_e}$ | 0.0633 | $C_{lp}$ | - |
| $C_{m0}$ | 0.0302 | $C_{lr}$ | - |
| $C_{m\alpha}$ | -0.126 | $C_{n0}$ | -0.000471 |
| $C_{m\delta_e}$ | -0.206 | $C_{n\beta}$ | 0.0283 |
| $C_{mq}$ | - | $C_{n\delta_a}$ | -0.00339 |
| $C_{C0}$ | 0.00316 | $C_{np}$ | - |
| $C_{C\beta}$ | -0.224 | $C_{nr}$ | - |

Fig. 15. shows the flight trajectory, including the 3D spatial position and airspeed variations. Fig. 16. presents the wind estimation results from the flight test. Subplots (a), (b), and (c) show the errors in the estimated speed of the horizontal wind, the estimated direction of the horizontal wind, and the estimated speed of the vertical wind. The black curve represents the reference value. The blue and red curves represent the differences between the wind speed estimated using the EKF and EKF+AMAE methods,

and the wind directly calculated using onboard sensor data, including angle of attack and sideslip angle.

The RMSE results are presented in Table XIV. The RMSE of the estimated horizontal wind speed using the EKF method is $1.9484 \text{ m/s}$, the RMSE of the horizontal wind direction is $10.9084°$, and the RMSE of the vertical wind speed is $0.7226 \text{ m/s}$. The RMSE of the estimated horizontal wind speed using the EKF+AMAE method is $1.8622 \text{ m/s}$, the RMSE of the horizontal wind direction is $9.7506°$, and the RMSE of the vertical wind speed is $0.8434 \text{ m/s}$. The RMSEs for wind speed estimation using both methods are within $2 \text{ m/s}$, and the estimation error for horizontal wind direction is around $10°$.

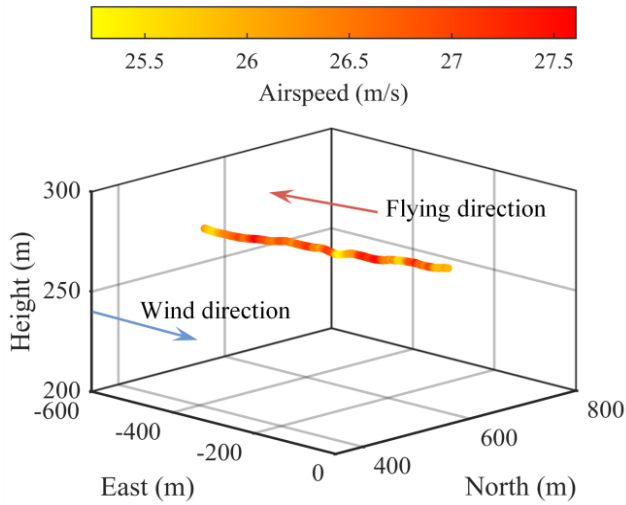

Fig. 15. Flight trajectory.

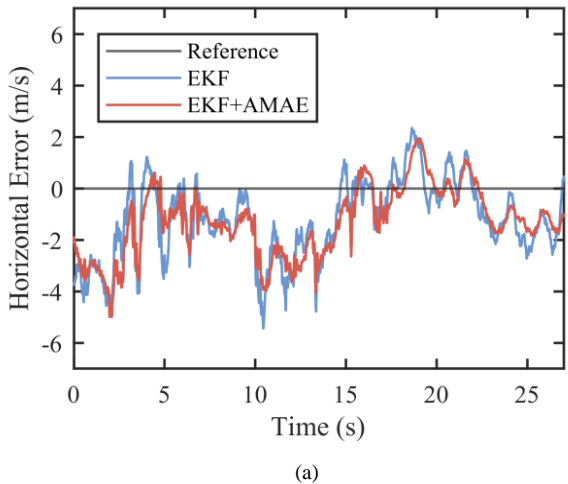

(a)

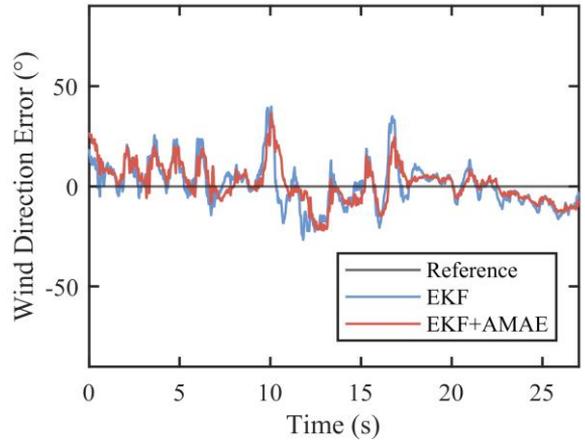

(b)

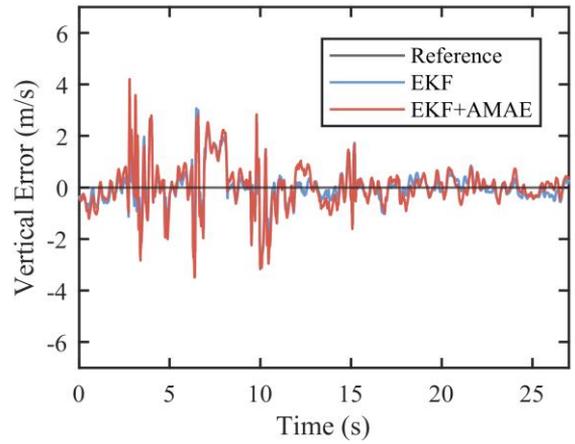

(c)

Fig. 16. Flight test estimation. (a) Error of horizontal wind speed. (b) Error of horizontal wind direction. (c) Error of vertical wind speed.

**TABLE XIV**
**RMSE of Flight Test Estimation**

| RMSE | EKF | EKF+AMAE |
|---|---|---|
| Horizontal wind (m/s) | 1.9489 | 1.8622 |
| Horizontal wind (°) | 10.9084 | 9.7506 |
| Vertical wind (m/s) | 0.7226 | 0.8434 |

The analysis of experimental conditions reveals the following causes of errors:

(1) Inaccuracies in the aerodynamic model affect estimation results. One reason for model inaccuracy is the use of a linear approximation model for wind estimation, which neglects nonlinear terms. Additionally, discrepancies exist between the aerodynamic model in Table XIII from the referenced study and the actual aerodynamic characteristics of the UAV during flight tests.

These discrepancies arise from platform modifications, airframe deformation under strong winds, and other factors.

(2) The lack of real wind data impacts estimation results. In simulations, wind can be predefined as reference values. However, during flight tests, no direct source of ground-truth wind information exists. This study evaluates errors by comparing estimates against calculated values (treated as references). Strong noise disturbances in onboard sensor data reduce reference value accuracy, causing significant deviations from true values and consequently larger estimation errors.

(3) Severe environmental disturbances introduce estimation errors. Actual flight tests encounter strong turbulent flows. Under current experimental conditions for small UAVs, low-cost sensors cannot accurately capture the instantaneous effects of turbulence on UAV dynamics, leading to substantial estimation errors.

Considering the above factors, the observed error level in real flight tests is deemed reasonable. As a low-cost wind estimation solution, the derived wind information can serve as a reference for UAVs to enhance autonomous environmental awareness. The analysis validates the method's effectiveness for wind estimation, while real-time performance during flight tests confirms its feasibility for onboard computation.

## V. CONCLUSION

This study proposes a wind estimation method for small UAVs that integrates an aerodynamic model with the EKF algorithm and introduces the AMAE method to achieve lower errors and smoother estimates. The approach efficiently estimates UAV states and wind using only essential onboard sensors (INS, GNSS, ADS) without requiring additional anemometry equipment.

Based on simulation analysis, this method confines wind speed estimation errors within $0.2 \text{ m/s}$ and wind direction estimation errors within $2°$, demonstrating its capability for low-cost, high-precision wind estimation. The study identified that inaccuracies in the aerodynamic model contribute to wind estimation errors, prompting a focused analysis on how model precision affects estimation accuracy. This provides researchers clearer insights into the error levels of estimated wind. Flight tests were conducted, showing horizontal wind speed estimation errors not exceeding $2 \text{ m/s}$, horizontal wind direction estimation errors around $10°$, and vertical wind speed estimation errors within $1 \text{m/s}$. Offline processing the test data of $27\text{s}$ on an arm platform took only $1.15\text{s}$, verifying both the method's effectiveness and its potential for onboard real-time computation. Finally, a comprehensive analysis of uncertainties and error sources in the tests was performed. In summary, the method's estimation accuracy and computational efficiency meet the requirements for autonomous wind estimation tasks on small UAVs.

Future research could further explore approaches to mitigate the impact of aerodynamic model inaccuracies on estimation results in low-cost wind estimation methods. Additionally, the proposed method could be applied to studies on wind energy utilization for energy-efficient autonomous flight of UAVs.

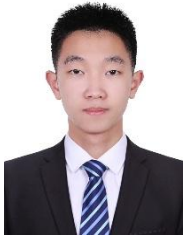

**Bingchen CHENG** was born in Hebei, China, in December 2003. He received the B.S. degree in aircraft design and engineering from Beihang University, Beijing, China, in June 2025. He is currently pursuing the Ph.D. degree in aircraft conceptual design and systems engineering with Beihang University. His research interests include intelligent flight control, trajectory planning, and multi-UAV swarm technology.

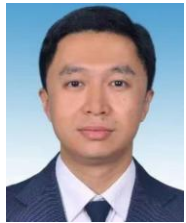

**Tielin MA** received the Ph.D. degree from Beihang University, Beijing, China, in 2008. He is currently working as a Researcher with Beihang University. His research interests include multidisciplinary optimization design of aircraft and overall design of unmanned aerial vehicles.

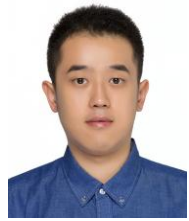

**Jingcheng FU** received the Ph.D. degree from Beihang University, Beijing, China, in 2021. He is currently working as an Associate Researcher with Beihang University. His research interests include intelligent flight control, unmanned aerial vehicle design, swarm UAV sensing.

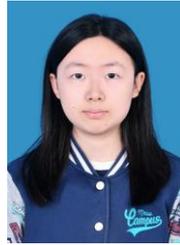

**Lulu TAO** was born in Guizhou, China, in February 2003. She received the B.S. degree in aircraft design and engineering from Beihang University, Beijing, China, in June 2025. She is currently pursuing a master's. degree in Unmanned Systems Science and Technology at Beihang University. Her research interests include aircraft system identification and intelligent flight control.

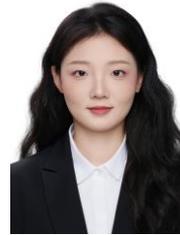

**Tianhui GUO** was born in Shannxi, China, in March 2001. She received the B.S. degree in aircraft design and engineering from Northwestern Polytechnical University, Shannxi, China, in June 2023. She is currently pursuing a master's. degree in aircraft design at Beihang University, Beijing, China. Her research interests include intelligent flight perception and control.